\documentclass[conference]{IEEEtran}
\IEEEoverridecommandlockouts
\usepackage{algorithm}
\usepackage{algpseudocode}
\usepackage{subcaption}
\usepackage{array}
\usepackage{booktabs}
\usepackage{longtable}
\usepackage{multicol}
\usepackage{multirow}
\usepackage{hyperref}
\usepackage{cite}
\usepackage{amsmath,amssymb,amsfonts}
\usepackage{graphicx}
\usepackage{textcomp}
\usepackage{xcolor}
\def\BibTeX{{\rm B\kern-.05em{\sc i\kern-.025em b}\kern-.08em
    T\kern-.1667em\lower.7ex\hbox{E}\kern-.125emX}}
\begin{document}

\title{A Compression Based Classification Framework Using Symbolic Dynamics of Chaotic Maps\\
}

\author{
\IEEEauthorblockN{Parth Naik\IEEEauthorrefmark{1}, Harikrishnan N. B.\IEEEauthorrefmark{1}\IEEEauthorrefmark{2}}
\IEEEauthorblockA{\IEEEauthorrefmark{1}Department of Computer Science and Information Systems (CSIS),\\
BITS Pilani, K. K. Birla Goa Campus, 403726, Goa, India}
\IEEEauthorblockA{\IEEEauthorrefmark{2}Adjunct Faculty, Consciousness Studies Programme,\\
National Institute of Advanced Studies, IISc Campus, Bengaluru, 560012, Karnataka, India}
\IEEEauthorblockA{Email: f20220491@goa.bits-pilani.ac.in, harikrishnannb@goa.bits-pilani.ac.in}
}

% \author{\IEEEauthorblockN{Parth Naik$^{a}$, Harikrishnan N B$^{a,b}$}
% \IEEEauthorblockA{\textit{$^{a}$Computer Science and Information Systems (CSIS)} \\
% \textit{BITS Pilani K K Birla Goa Campus, Goa, India}\\
% \textit{Consciousness Studies Programme\\ National Institute of Advanced Studies\\IISc Campus, Bengaluru, India}\\
% f20220491@goa.bits-pilani.ac.in, harikrishnannb@goa.bits-pilani.ac.in}
% \and
% \IEEEauthorblockN{Harikrishnan N B}
% \IEEEauthorblockA{\textit{Computer Science and Information Systems (CSIS)} \\
% \textit{BITS Pilani K K Birla Goa Campus}\\
% Goa, India \\
% harikrishnannb@goa.bits-pilani.ac.in}
% \and
% \and
% \and
% \and

\maketitle

\begin{abstract}
We propose a novel classification framework grounded in symbolic dynamics and data compression using chaotic maps. The core idea is to model each class by generating symbolic sequences from thresholded real-valued training data, which are then evolved through a one-dimensional chaotic map. For each class, we compute the transition probabilities of symbolic patterns (e.g., `00', `01', `10', and `11' for the second return map) and aggregate these statistics to form a class-specific probabilistic model. During testing phase, the test data are thresholded and symbolized, and then encoded using the class-wise symbolic statistics via back iteration, a dynamical reconstruction technique. The predicted label corresponds to the class yielding the shortest compressed representation, signifying the most efficient symbolic encoding under its respective chaotic model. This approach fuses concepts from dynamical systems, symbolic representations, and compression-based learning. We evaluate the proposed method: \emph{ChaosComp} on both synthetic and real-world datasets, demonstrating competitive performance compared to traditional machine learning algorithms (e.g., macro F1-scores for the proposed method on Breast Cancer Wisconsin = 0.9531, Seeds = 0.9475, Iris = 0.8469 etc.). Rather than aiming for state-of-the-art performance, the goal of this research is to reinterpret the classification problem through the lens of dynamical systems and compression, which are foundational perspectives in learning theory and information processing.
\end{abstract}


\section{Introduction}
Machine learning can be understood through several foundational perspectives, each offering unique insights into what it means for a system to “learn” from data. One classical perspective views learning as a problem of function approximation. Here, the goal is to learn a function \( f : X \rightarrow Y \) that maps input features to output labels or values. This function tries to minimize the error between the predicted value and the actual value. To measure this error, a loss function is used. The loss function gives a numerical value that shows how far the prediction is from the actual output. The goal of the learning algorithm is to find the function that gives the smallest total loss on the training data. This method is used in most of the supervised learning models like linear regression, support vector machines, decision trees and neural networks. 

From the perspective of probabilistic inference, the process of learning is fundamentally about identifying the most probable explanation for observed data. This framework goes beyond simply establishing a function that associates inputs with outputs; it emphasizes the importance of grasping the underlying uncertainty and distribution inherent in the data. A common way to do this is through Maximum Likelihood Estimation (MLE), which finds the parameters that make the observed data most probable. Another method is Maximum A Posteriori (MAP) estimation, which also considers prior knowledge or beliefs about the parameters. This perspective is the basis for many popular algorithms, such as Naïve Bayes, logistic regression, Hidden Markov Models (HMMs), and Bayesian networks.

Another important and less conventional way to understand learning is through the lens of data compression, which is formalized by the Minimum Description Length (MDL) principle. This approach mainly focuses on learning the data using the least discription possible. The central idea is that a good model should not only fit the data well but also be simple enough to avoid overfitting. In other words, the best model is the one that provides the shortest total description of itself and the data when encoded using that model~\cite{grunwald2007mdl}.

The MDL principle has its roots in information theory and algorithmic complexity, particularly in the concepts of Kolmogorov complexity and Shannon entropy. According to MDL, learning from data is equivalent to selecting the shortest description that best explains it. This scheme should be capable of explaining the data using the fewest number of bits. In practice, the total description length is often divided into two parts: one for encoding the model (or hypothesis) and one for encoding the data given that model. This leads to the classical MDL formulation:
\begin{equation}
\text{Total Description Length} = L(h) + L(D \mid h)
\end{equation}

where \( L(h) \) is the length of the model description and \( L(D \mid h) \) is the length of the data encoded using the model \( h \)~\cite{vitanyi_learning_as_compression}.

One of the key strengths of the MDL perspective is that it naturally balances model complexity and data fit, without needing explicit regularization terms or prior distributions as in probabilistic methods. This makes MDL particularly attractive in cases where we lack reliable prior information or want to avoid making strong probabilistic assumptions. MDL has been successfully applied to various tasks such as classification, model selection, clustering, and even sequence prediction~\cite{adriaans2008learning, grunwald2007mdl}.

Although the ideal MDL approach is based on Kolmogorov complexity, which is uncomputable in general, practical approximations exist. For classification, our method calculates how well different classes can compress the data and choose the model that results in the shortest file size. This idea has been explored in algorithms like Prediction by Partial Matching (PPM) and compression-based clustering~\cite{marton2005compression,cilibrasi2005clustering}.

Overall, viewing learning as data compression emphasizes that a good model is one that captures meaningful structure in the data. The more a model compresses, the more it has “understood” the data. This perspective has been explored using symbolic sequence modeling from chaotic maps—for instance, using second-order skew tent maps for classification \cite{vats2025chaotic}. The Baker's map, Tent map, and its skewed variant are known to be ergodic and exhibit robust chaos~\cite{lawnik2025transformation}, characterized by a positive Lyapunov exponent~\cite{alligood1997chaos} across all values of the skewness parameter. These chaotic maps have found diverse applications, including pseudorandom number generation in cryptography~\cite{lawnik2025transformation}, joint image compression and encryption~\cite{tong2017joint}, and classification tasks~\cite{harikrishnan2021noise}. Notably, it has been shown that piecewise linear chaotic maps are Shannon-optimal for source coding~\cite{NAGARAJ20091013}, a result that forms a key theoretical foundation for the current research.
In this work, we extend this idea by employing the Baker's map and higher-order symbolic encoding to design a more flexible compression-based classifier.

\section{Background}
This section introduces the foundational concepts underpinning our algorithm. We begin with the formal definition of the Baker's map, then describe the back‐iteration procedure used to recover the interval of initial conditions corresponding to a given symbolic sequence. Next, we review the proof of the Baker's map’s Shannon‐optimal coding properties and finally extend these ideas to the general $n$-th return map.

\subsection{Baker's Map}
\begin{figure}[h]
    \centering
    \includegraphics[width=0.6\linewidth]{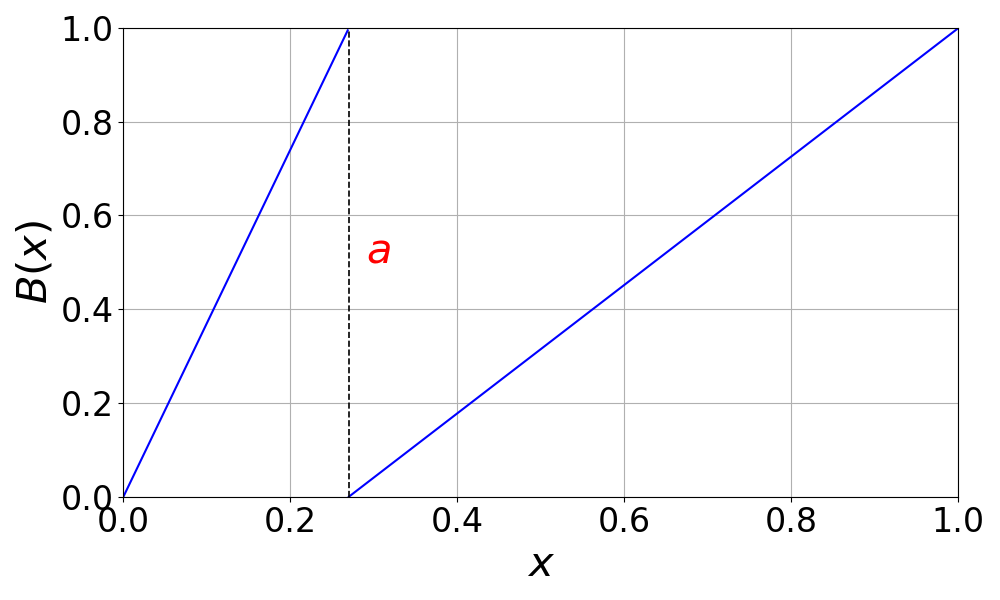}
    \caption{Baker's map with a = 0.27}
    \label{fig:baker_map}
\end{figure}

Baker's maps are chaotic, piecewise linear maps. The mathematical definition of the map is given by equation \ref{baker_eq}: 

\begin{equation}
\label{baker_eq}
  B(x) = 
  \begin{cases}
    \frac{x}{a} & \text{for } 0 \leq x < a \\
    \frac{x - a}{1 - a} & \text{for } a \leq x < 1
  \end{cases}
\end{equation}
    where \( a \in (0, 1)\) determines the skewness of the Baker's map. An example of a Baker's map with a = 0.27 is shown in Fig~\ref{fig:baker_map}.

Given a real-valued initial condition \( x_0 \in (0,1) \), the Baker's map is iteratively applied to generate a chaotic trajectory. Let the trajectory generated by forward iteration of the Baker's map be denoted as:
\[
x_0 \rightarrow x_1 \rightarrow x_2 \rightarrow \ldots \rightarrow x_n \rightarrow \ldots
\]
This trajectory depends on the initial value \( x_0 \) and a fixed parameter \( a \in (0,1) \) of the Baker's map. The real-valued sequence \( \{x_i\} \) can be converted into a symbolic sequence \( \{s_i\} \) using the thresholding rule:

\begin{equation}
    \label{sym_seq}
    s_i(x_i) = 
    \begin{cases}
        0 & \text{if } x_i < a, \\
        1 & \text{if } x_i \geq a.
    \end{cases}
\end{equation}

\noindent \textbf{Example:} Let \( a = 0.4 \) and \( x_0 = 0.3 \). Then:
\begin{align*}
x_0 &= 0.3 < a \quad &&\Rightarrow s_0 = 0, \\
x_1 &= \frac{x_0}{a} = \frac{0.3}{0.4} = 0.75 > a \quad &&\Rightarrow s_1 = 1, \\
x_2 &= \frac{x_1 - a}{1 - a} = \frac{0.75 - 0.4}{0.6} = 0.5833\ldots > a \quad &&\Rightarrow s_2 = 1, \\
x_3 &= B(x_2) \quad &&\text{and so on.}
\end{align*}

\noindent Hence, the symbolic sequence is \( S = 0\,1\,1\,\ldots \)

% Despite its chaotic nature, the Baker's map is piecewise invertible. Inverse mapping $B^{-1}$ reconstructs the previous states of the system based on the symbolic sequence $S$:

% \begin{equation}
%     B^{-1}(x_{t+1}, s_t) =
%     \begin{cases}
%     a x_{t+1}, &\text{if } s_t = 0, \\
%     (1 - a)x_{t+1} + a, &\text{if } s_t = 1.
%     \end{cases}
% \end{equation}

% Given a symbolic sequence $s_T, s_{T-1}, \ldots, s_1$ and a terminal state $x_T \in [0,1]$ (e.g., $x_T = 0.5$), one can iteratively compute:
% \[
% x_{t} = B^{-1}(x_{t+1}, s_t), \quad t = T-1, T-2, \ldots, 0.
% \]

% In our backward iteration method, only the final symbolic value $s_T$ is known initially. Based on $s_T$, the corresponding subinterval is identified, and its bounds are used to initialize the reconstruction process. These bounds are then iteratively propagated backward using the inverse of the Baker's map, with each preceding symbol $s_{T-1}, s_{T-2}, \ldots, s_1$ guiding the selection and transformation of the interval at each step. This process yields an approximation of the original input $x_0$ as shown in algorithm \ref{alg:baker_backward} by tracing the symbolic trajectory in reverse.

\subsection{Inverse Problem: Symbolic Sequence to Initial Interval}

Given a symbolic sequence \( S = s_0, s_1, \ldots, s_T \in \{0,1\}^{T+1} \) and a fixed Baker's map parameter \( a \in (0,1) \), we pose the following inverse problem:

\begin{quote}
\textbf{Problem:} Find the set of all initial values \( x_0 \in (0,1) \) such that forward iteration under the Baker's map \( B(\cdot) \) starting from the initial value $x_0$, followed by threshold-based symbolic encoding (defined in Equation~\ref{sym_seq}), produces the symbolic sequence \( s_0, s_1, \ldots, s_T \).
\end{quote}

Despite its chaotic behavior, the Baker's map is piecewise linear and invertible on its subdomains. This structure allows us to approach the inverse problem via backward interval iteration using the symbolic sequence. 

We define the inverse branches of the Baker's map \( B^{-1} \) as follows:
\begin{equation}
    B^{-1}(x_{t+1}, s_t) =
    \begin{cases}
    a \cdot x_{t+1}, & \text{if } s_t = 0, \\
    (1 - a) \cdot x_{t+1} + a, & \text{if } s_t = 1.
    \end{cases}
\end{equation}

\noindent Given the final symbol \( s_T \), we define the corresponding subinterval \( I_T \subset (0,1) \) as:
\begin{equation}
    I_T = 
    \begin{cases}
    [0, a), & \text{if } s_T = 0, \\
    [a, 1), & \text{if } s_T = 1.
    \end{cases}
\end{equation}

\noindent We then iteratively propagate this interval backwards using the symbolic sequence:
\begin{equation}
    I_t = 
    \begin{cases}
    a \cdot I_{t+1}, & \text{if } s_t = 0, \\
    (1 - a) \cdot I_{t+1} + a, & \text{if } s_t = 1,
    \end{cases}
    \quad \text{for } t = T-1,\ldots, 0.
\end{equation}

\noindent Here, scalar multiplication of an interval denotes scaling both endpoints. That is, for an interval \( I = [L, U] \), we define:
\[
a \cdot I = [a \cdot L, a \cdot U], \quad
(1 - a) \cdot I + a = [(1 - a) \cdot L + a, (1 - a) \cdot U + a].
\]

After \( T+1 \) steps, the interval \( I_0 =[L, U] \subset (0,1) \) contains all possible values of \( x_0 \) such that forward iteration under the Baker's map (with symbolic thresholding at \( a \)) yields the exact symbolic sequence \( s_0, s_1, \ldots, s_T \). This method guarantees that:
\[
\forall x_0 \in I_0, \quad s_i(x_i) = s_i \quad \text{for } i = 0, 1, \ldots, T.
\]

\noindent As \( T \to \infty \), the length of \( I_0 \) shrinks exponentially, converging to a unique point if the sequence is infinitely long.
\subsection{Compressed File Size from Interval Length}

The length of the interval $\ell = U - L$ represents the probability assigned to the symbolic sequence $S$ under the model parameterized by $a$ (assuming independent and identical distribution). This interpretation is grounded in the framework of arithmetic coding, where symbolic sequences are mapped to subintervals of $[0,1)$ and the interval length corresponds to their probability.

According to Shannon's source coding theorem, the optimal number of bits required to encode a message (symbolic sequence) ith probability $\ell$ is:
\begin{equation}
    \text{Compressed file size} = \lceil-\log_2(\ell) \rceil= \lceil-\log_2(U - L)\rceil.
\end{equation}

This quantity is interpreted as the \emph{description length}. This relationship between symbolic dynamics, interval length, and information content has been formally established in \cite{NAGARAJ20091013}, where it is shown that chaotic maps like the skewed Tent map can be used to implement Shannon-optimal arithmetic coding schemes.

\subsection{Optimal Choice of Parameter $a$}
To ensure minimal description length (i.e., optimal compression), the threshold parameter $a$ should match the empirical probability of observing a `0' in the symbolic sequence $S = \{s_0,s_1,s_2,\ldots,..s_T\}$. That is,
\[
a = \frac{\text{\# of 0s in } S}{T+1}.
\]
where $T+1$ is the length of the symbolic sequence $S$. This choice of $a$ ensures that the piece-wise linear map map-based coding achieves \emph{Shannon optimality}, as demonstrated in \cite{NAGARAJ20091013}. This means that the final interval produced via back-iteration has length equal to the true probability of the symbolic sequence, allowing the symbolic sequence to be compressed to its entropy rate.

\begin{algorithm}[h]
\caption{Backward Iteration}
\label{alg:baker_backward}
\begin{algorithmic}[1]
\Require Symbolic sequence $S = (s_0, s_1, \ldots, s_n)$, parameter $a$
\Ensure Approximate origin $x_0$
\State $S' \gets$ reverse$(S)$
\If{$S'_0 = 0$}
    \State $L \gets 0$, $U \gets a$
\Else
    \State $L \gets a$, $U \gets 1$
\EndIf
\For{$i = 1$ to $n$}
    \If{$S'_i = 0$}
        \State $L \gets a \cdot L$
        \State $U \gets a \cdot U$
    \Else
        \State $L \gets (1 - a) \cdot L + a$
        \State $U \gets (1 - a) \cdot U + a$
    \EndIf
    \If{$L > U$}
        \State swap$(L, U)$
    \EndIf
\EndFor
\State \Return $x_0 \gets \frac{L + U}{2}$ \textbf{Interval} L,U
\end{algorithmic}
\end{algorithm}

\subsection{Shannon Optimality of Baker's Map Coding}

The proof for Shannon optimality for skew tent maps is provided in~\cite{NAGARAJ20091013}. We demonstrate that the Baker's map, when used for symbolic compression, achieves Shannon’s entropy rate for binary independent and identically distributed (i.i.d.) sources.

Consider a binary i.i.d. sequence that has two symbols, 0 and 1, occurring with a probability of $p$ and $(1-p)$ respectively. The Shannon entropy rate of such a source is given by:
\begin{equation}
    H = -p \log_2(p) - (1 - p) \log_2(1 - p) \quad \text{(bits/symbol)}.
\end{equation}

Consider a binary symbolic sequence \( S\) of length \( N \), consisting of \( k \) symbols \( 0 \) and \( N - k \) symbols \( 1 \). When $ N \rightarrow \infty $, we have $\frac{k}{N} \rightarrow p$ The probability of observing such a symbolic sequence is:
\begin{equation}
    P(S) = \Big(\frac{k}{N}\Big)^k \Big(1 - \frac{k}{N}\Big)^{N - k}.
\end{equation}

In Baker's map coding, we interpret each binary symbol as a transformation on the unit interval \([0, 1]\). The interval shrinks based on the symbol:
\begin{itemize}
    \item If the symbol is \( 0 \), the interval is shrunk by a factor of $\frac{k}{N}$.
    \item If the symbol is \( 1 \), the interval is shrunk by a factor of $\Big(1 - \frac{k}{N}\Big)$.
\end{itemize}

The length of the final interval becomes:
\begin{equation}
    \text{Length} = \Big(\frac{k}{N}\Big)^k \Big(1 - \frac{k}{N}\Big)^{N - k} = P(S).
\end{equation}

To code the initial condition in the interval, we need 
\begin{equation}
    \left\lceil -\log_2(\text{Length}) \right\rceil = \left\lceil -\log_2(P(S)) \right\rceil \quad \text{bits}.
\end{equation}

Expanding the equation gives us:
\begin{equation}
    \leq -k \log_2\Big(\frac{k}{N}\Big) - (N-k) \log_2\Big(1-\frac{k}{N}\Big) + 1 
\end{equation}

We divide the expression by N to get the number of bits per symbol in the compressed file:
{\small
\begin{equation}
\begin{aligned}
    % \frac{1}{N}  \left\lceil -\log_2(\text{Length}) \right\rceil  
    \leq -\Big(\frac{k}{N}\Big) \log_2\Big(\frac{k}{N}\Big) - \Big(1-\frac{k}{N}\Big) \log_2\Big(1-\frac{k}{N}\Big) + \frac{1}{N} \\
    = -p \log_2(p) - (1 - p) \log_2(1 - p) + \frac{1}{N} = H + \frac{1}{N}.
\end{aligned}
\end{equation}
}
When $N \rightarrow \infty$ this expression tends to $H$. Thus Baker's map coding achieves Shannon’s entropy rate of the source as the length of the symbolic sequence goes to infinity.

\subsection{Second-Return Map and Pairwise Symbolic Coding}

In the standard (first-return) formulation, backward iteration is performed using one symbol at a time (e.g., $s_0, s_1, \ldots,s_T$) and a single-parameter map (e.g., the Baker's map with parameter $a$). However, this can be extended to higher-order symbolic dynamics for improved modeling fidelity. In particular, we consider the \textbf{second-return map}, where the symbolic sequence $S=s_0, s_1, \ldots,s_T$ is grouped into non-overlapping pairs of bits:  
\[
(s_0, s_1), (s_2, s_3), \ldots, (s_{T-1}, s_T).
\]

assuming \( T \) is even. Each pair belongs to the set \( \mathcal{S}_2 = \{00, 01, 10, 11\} \).

Let \( N_{ij} \) denote the number of times the symbol pair \( ij \in \mathcal{S}_2 \) appears in the sequence. The empirical probability of each pair is then computed as:

\begin{equation}
    p_{ij} = \frac{N_{ij}}{\sum_{kl \in \mathcal{S}_2} N_{kl}},
    \label{eq:pairwise-prob}
\end{equation}

\noindent where \( \sum_{ij} p_{ij} = 1 \), provided all counts are non-zero. These probabilities define a partition of the unit interval \([0,1)\) into four subintervals of lengths \( p_{00}, p_{01}, p_{10}, p_{11} \), respectively.

\subsubsection*{Defining the Second-Return Map}

Using the pairwise probabilities from Eq.~\eqref{eq:pairwise-prob}, we define a piecewise-linear chaotic map \( B^{(2)}: [0,1) \to [0,1) \) as:

\small{
\begin{equation}
    B^{(2)}(x) = 
    \begin{cases}
        \dfrac{x}{p_{00}}, \quad\text{for } 0 \leq x < p_{00}, \\[6pt]
        \dfrac{x - p_{00}}{p_{01}},\quad\text{for } p_{00} \leq x < p_{00} + p_{01}, \\[6pt]
        \dfrac{x - (p_{00} + p_{01})}{p_{10}},  \quad\text{for } p_{00} + p_{01} \leq x < 1- p_{11}, \\[6pt]
        \dfrac{x - (p_{00} + p_{01} + p_{10})}{p_{11}},\quad \text{for } 1 - p_{11} \leq x < 1.
    \end{cases}
    \label{eq:second-return-map}
\end{equation}}

Each subinterval is mapped linearly back to the full interval \([0,1)\), ensuring ergodicity, and chaotic behavior. This map encodes symbolic dynamics based on pairwise statistics extracted directly from the symbolic sequence. Figure~\ref{fig:2-order_map} depicts graphical representation of second return map with empirical probabilities.
\subsection{Generalization to the $n$-th return Map}

The construction of the second-return map can be naturally extended to symbolic blocks of length \( n \). Given a symbolic sequence \( S = s_0, s_1, \ldots, s_T \), we extract non-overlapping subsequences of length \( n \):
\[
(s_0, \ldots, s_{n-1}),\ (s_n, \ldots, s_{2n-1}),\ \ldots,\ (s_{T-n+1}, \ldots, s_T),
\]
which we denote as elements of the set \( \mathcal{S}_n = \{0,1\}^n \), the collection of all binary sequences of length \( n \).

Let \( N_{w} \) denote the number of occurrences of an \( n \)-length word \( w \in \mathcal{S}_n \) in the symbolic sequence. The empirical probability of each word is defined as:
\begin{equation}
    p_{w} = \frac{N_{w}}{\sum_{u \in \mathcal{S}_n} N_{u}}, \quad \text{for each } w \in \mathcal{S}_n.
    \label{eq:nth-symbol-prob}
\end{equation}

\subsubsection*{Definition of the $n$-th Return Map}

Using the empirical distribution \( \{p_w\}_{w \in \mathcal{S}_n} \), we partition the unit interval \([0,1)\) into \( 2^n \) subintervals, each of length \( p_w \). For notational simplicity, assume the elements of \( \mathcal{S}_n \) are indexed lexicographically as \( w_0, w_1, \ldots, w_{2^n-1} \), with corresponding probabilities \( p_0, p_1, \ldots, p_{2^n-1} \).

The symbolic dynamics is then defined via a piecewise-linear map \( B^{(n)}(x) \), which maps each subinterval linearly onto the full interval \([0,1)\), as follows:

\begin{equation}
\label{n-ordered_eq}
    B^{(n)}(x) = 
    \begin{cases}
        \frac{x}{p_{0}} &\quad \text{for } 0 \leq x < p_{0}, \\[1em]
        \frac{x - p_{0}}{p_{1}} &\quad \text{for } p_{0} \leq x < p_{0} + p_{1}, \\[1em]
        \vdots & \\[0.5em]
        \frac{x - \sum_{j=0}^{2^{n}-2} p_j}{p_{2^{n}-1}} &\quad \text{for } \sum_{j=0}^{2^{n}-2} p_j \leq x < 1.
    \end{cases}
\end{equation}

As in the second-return case, the map \( B^{(n)} \) is expanding, piecewise-linear, and onto. It defines symbolic dynamics over blocks of length \( n \), enabling both encoding via forward iteration and symbolic reconstruction via backward iteration.

\section{Methodology}
 
Consider a dataset which contains $m$ classes, with each data sample containing $k$ features. To ensure consistency, we apply MinMax scalar \cite{scikit-learn}, mapping each feature to the interval [0,1]. This dataset is then split into training and testing sets. Symbolic sequences are achieved by binarizing the training data using a threshold. The thresholding process is explained in a further section. From these symbolic sequences, non-overlapping subsequences of length $n$ are extracted. However, some features are left out of the feature vector, since the number of features i.e $k$ is not always a multiple of $n$. We extend the feature vector by applying a padding strategy to guarantee complete coverage of original features. Adding dummy feature values to every data instance is known as padding.
For instance, consider a symbolic sequence $S$ = [1,0,0,1], where $k$ = 4. For $n$ = 3, the last feature will be ignored if we do not pad the feature vector. To make the total length a multiple of $n$, we need to pad $n$-($k$\%$n$) features to it. In this case, $n$-($k$\%$n$) is 2, hence two dummy symbols (either 0 or 1) are appended. The padded sequence that is produced is $S$ =[1,0,0,1,1,1]. To ensure uniformity, this padding procedure is applied consistently across all data instances of all $m$ classes in the symbolic encoding process. 

Given a symbolic sequence length parameter \( n \), there are \( 2^n \) possible binary subsequences of length \( n \), each representing a symbolic pattern that may occur within the feature space. The distribution of these subsequences—i.e., their empirical frequencies determines the skewness and structure of the resulting \(n^{\text{th}}\) return map, which is central to the symbolic encoding process.

% To illustrate, consider a dataset with \( m = 2 \) classes, \( k = 4 \) features per instance, and \( n = 2 \). Each class contains three instances. The possible symbolic subsequences for \( n = 2 \) are: \((00), (01), (10), (11)\). We extract all non-overlapping subsequences of length \( n \) from each data instance. For Class 0, with samples \([0\ 1\ 1\ 0], [1\ 1\ 1\ 0], [0\ 0\ 1\ 0]\), the extracted subsequences are:
% \[
% (01), (10), (11), (10), (00), (10)
% \]
% For Class 1, with samples \([0\ 0\ 0\ 1], [1\ 0\ 1\ 0], [1\ 1\ 0\ 0]\), the extracted subsequences are:
% \[
% (00), (01), (10), (10), (11), (00)
% \]

% Based on these, we compute the empirical probability distribution of each subsequence for every class. For example:
% \begin{itemize}
%     \item {Class 0}: \( p_{00} = \frac{1}{6},\ p_{01} = \frac{1}{6},\ p_{10} = \frac{3}{6},\ p_{11} = \frac{1}{6} \)
%     \item {Class 1}: \( p_{00} = \frac{2}{6},\ p_{01} = \frac{1}{6},\ p_{10} = \frac{2}{6},\ p_{11} = \frac{1}{6} \)
% \end{itemize}

% These class-specific probability distributions define the structure of their respective n-th return map. Figure~\ref{fig:2-order_map} illustrates the second return map corresponding to the probability distribution of Class 1. This symbolic representation allows us to encode class-specific sequence patterns that can be leveraged for classification by our classifier: ChaosComp.

To illustrate the construction of class-specific symbolic models, consider a dataset with \( m = 2 \) classes, where each data instance consists of \( k = 4 \) binary features. Let the symbolic block length be \( n = 2 \), corresponding to a second-return map. The set of possible symbolic subsequences of length \( n \) is \( \mathcal{S}_2 = \{00, 01, 10, 11\} \).

We extract all non-overlapping 2-length subsequences from each data instance. For example:

\begin{itemize}
    \item \textbf{Class 0}: Suppose it contains the binary sequences \([0\ 1\ 1\ 0]\), \([1\ 1\ 1\ 0]\), and \([0\ 0\ 1\ 0]\). The extracted subsequences are:
    \[
    (01),\ (10),\ (11),\ (10),\ (00),\ (10)
    \]
    
    \item \textbf{Class 1}: With samples \([0\ 0\ 0\ 1]\), \([1\ 0\ 1\ 0]\), and \([1\ 1\ 0\ 0]\), the extracted subsequences are:
    \[
    (00),\ (01),\ (10),\ (10),\ (11),\ (00)
    \]
\end{itemize}

From these subsequences, we compute the empirical probability distribution over \( \mathcal{S}_2 \) for each class. For example:

\begin{itemize}
    \item \textbf{Class 0}:
    \[
    p_{00} = \frac{1}{6},\quad
    p_{01} = \frac{1}{6},\quad
    p_{10} = \frac{3}{6},\quad
    p_{11} = \frac{1}{6}
    \]
    
    \item \textbf{Class 1}:
    \[
    p_{00} = \frac{2}{6},\quad
    p_{01} = \frac{1}{6},\quad
    p_{10} = \frac{2}{6},\quad
    p_{11} = \frac{1}{6}
    \]
\end{itemize}

\begin{figure}[h]
    \centering
    \includegraphics[width=0.6\linewidth]{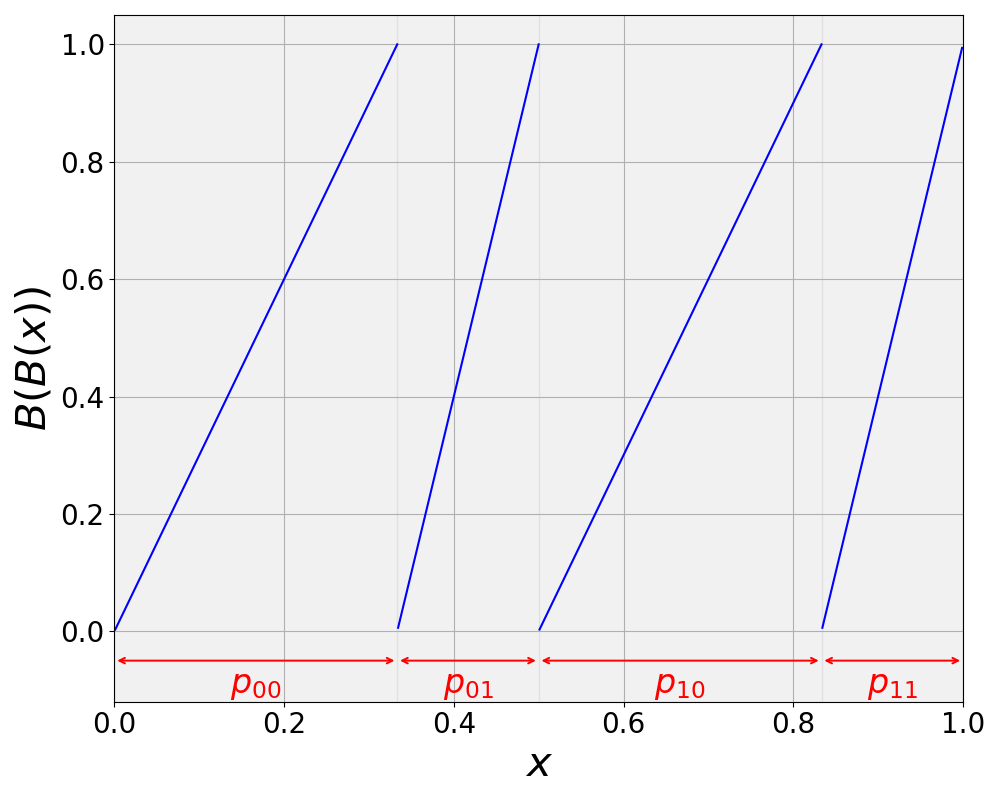}
    \caption{Second return Baker's map with empirical probabilities $p_{00}$ = $\frac{2}{6}$, $p_{01}$ = $\frac{1}{6}$, $p_{10}$ =  $\frac{2}{6}$, $p_{11}$ = $\frac{1}{6}$ for class 1.}
    \label{fig:2-order_map}
\end{figure}

These class-specific probability distributions define the structure of the corresponding second-return chaotic maps. Each probability \( p_{w} \), $w \in \mathcal{S}_2 $ determines the width of a subinterval in \([0,1)\), and the associated map stretches that subinterval linearly back to \([0,1)\). Figure~\ref{fig:2-order_map} shows the piecewise-linear map for Class 1 based on its empirical distribution. This symbolic representation captures class-specific sequential dependencies and forms the foundation for classification in our proposed classifier, \textit{ChaosComp}.

\begin{algorithm}
\caption{Backward Iteration for n-th Return map}
\label{alg:n_baker_backward}
\begin{algorithmic}[1]
\Require Symbolic sequence $S = (s_0, s_1, \ldots, s_n)$, parameters $p_{0},p_{1},p_{2},\ldots,p_{2^{n}-1}$
\Ensure Approximate origin $x_0$
\For{$i = 0$ to $2^{n}-1$}
\If{$S_0 = b_{i}$}
    \State $L \gets 0 + \sum_{j=0}^{i-1} p_{j}$, $U \gets 0 + \sum_{j=0}^{i} p_{j}$
\EndIf
\EndFor

\For{$i = 1$ to $n$}
    \For{$j = 0$ to $2^{n}-1$}
        \If{$S_i = b_{j}$}
            \State $L \gets p_{j} \cdot L$ + $\sum_{k=0}^{j-1} p_{k}$
            \State $U \gets p_{j} \cdot U$ + $\sum_{k=0}^{j-1} p_{k}$
        \EndIf
        \If{$L > U$}
            \State swap$(L, U)$
        \EndIf
    \EndFor    
\EndFor
\State \Return $x_0 \gets \frac{L + U}{2}$, \textbf{Interval} L,U
\end{algorithmic}
\end{algorithm}

\subsection{Preprocessing of dataset}
 \emph{ChaosComp} is applicable to datasets with atleast two features. Given a dataset with \( k \) features, we augment it by adding a new feature defined as the sum of the squares of the original \( k \) features, resulting in a total of \( k+1 \) dimensions. This augmentation is applied only when \( k < 30 \), as back iteration becomes numerically unstable for higher-dimensional data due to finite-precision limitations.\footnote{For \( k \geq 30 \), numerical issues in back iteration arise due to the shrinking interval size. This can potentially be addressed using renormalization techniques from arithmetic coding~\cite{sayood2017introduction}.} Subsequently, all features are normalized to the \([0,1]\) range using a MinMax scaler. 
 % In a separate set of experiments, Principal Component Analysis (PCA) is also employed as a preprocessing step to reduce dimensionality and evaluate the robustness of \emph{ChaosComp} under feature compression.

 \subsection{Hyperparameters for \emph{ChaosComp}}
 The preprocessed data is binarized using a threshold which a hyperparameter for \emph{ChaosComp}. Using steps of $0.01$, a threshold ranging from 0.01 to 1.00, is used to binarize the data. Values $\geq$ threshold are mapped to 1 otherwise 0. For the $n$-th return map, this thresholding converts the data into symbolic sequences. Padding is carried out when the number of features is not a multiple of $n$. The value of $n$ is also a hyperparameter. 
 % In experiments where PCA is used as a preprocessing step, $n\_components$ acts as a hyperparameter that determines the number of features to which the dataset is reduced.

\subsection{Training }
During training, for a dataset with \( m \) classes, \( m \) distinct \( n \)-th return Baker's maps are constructed—one per class. Each map estimates \( 2^n \) empirical probabilities corresponding to symbolic subsequences of length \( n \). Thus, the total number of learnable parameters in the model is \( m \times 2^n \).

For each training instance of a class, the empirical probabilities of all \( n \)-length binary subsequences are computed and then averaged across instances to obtain class-specific probability distributions.

If a particular subsequence never occurs in the training data of a class, its empirical probability becomes zero, which disrupts the backward iteration process. To address this, Laplace smoothing is applied to ensure non-zero probabilities:

\begin{equation}
\label{Laplace_smoothing}
\hat{p}_w = \frac{\sum_{i=1}^{N} p_{i,w} + \alpha}{N + 2^n \cdot \alpha}, \quad \text{for all } w \in \{0,1\}^n
\end{equation}

\noindent \textbf{where:}
\begin{itemize}
    \item \( w \) is an \( n \)-length binary subsequence.
    \item \( p_{i,w} \) is the empirical probability of \( w \) in the \( i^\text{th} \) training instance.
    \item \( N \) is the number of training instances in the class.
    \item \( \alpha \) is the Laplace smoothing parameter.
\end{itemize}
%%%%%%%%%
\subsection{Testing} 
During the testing phase, the algorithm employs the backward iteration (algorithm~\ref{alg:n_baker_backward}) procedure for classification. Each test instance is binarized using the optimal threshold determined during hyperparameter tuning. The resulting symbolic sequence, along with the empirical class-wise probability distributions, is input to the backward iteration method which returns the original input value and the interval $(L,U)$. For each class, the compressed file size is computed by:
\begin{equation}
    \text{Compressed file size} = \lceil -\log_{2}(U-L) \rceil
\end{equation}

The class yielding the smallest compressed file size is selected as the predicted label. In the event of a tie—where multiple classes produce the same file size—cosine similarity is used as a secondary criterion. For example, if the symbolic sequence for the test sample is [0 0 1 1], then the empirical probabilities for second Return map of this test sample will be $p_{00}$ = 0.5, $p_{01}$ = 0.0, $p_{10}$ = 0.0, $p_{11}$ = 0.5. After Laplace smoothing (with $\alpha$ = 0.01), the empirical probabilities will be $p_{00}$ = 0.4903, $p_{01}$ = 0.0097, $p_{10}$ = 0.0097, $p_{11}$ = 0.4903. Now these probabilities are treated as a vector, $[0.4903, 0.0097, 0.0097, 0.4903]$, and cosine similarity is calculated with the empirical probability vector of each class. The class with the highest cosine similarity is selected as the predicted label. 

% Modifying the template --- including but not limited to: adjusting
% margins, typeface sizes, line spacing, paragraph and list definitions,
% and the use of the \verb|\vspace| command to manually adjust the
% vertical spacing between elements of your work --- is not allowed.

% {\bfseries Your document will be returned to you for revision if
%   modifications are discovered.}
\section{Experiments and Results}
We conducted three sets of experiments to evaluate the performance of the proposed method:

\begin{itemize}
    \item \textbf{Experiment 1:} Training and testing on synthetic 2D datasets, including \textit{Circles}, \textit{Moons}, and \textit{Linearly Separable} data, to visually assess decision boundaries and model behavior.
    
    \item \textbf{Experiment 2:} Training and testing on standard real-world classification datasets, namely \textit{Iris}, \textit{Wine}, \textit{Breast Cancer Wisconsin}, \textit{Seeds}, \textit{Ionosphere}, and \textit{Banknote Authentication}.
\end{itemize}

\subsection{XOR problem}
To evaluate the capacity of our algorithm on non-linearly separable data, we applied it to the classical XOR classification problem. The XOR dataset consists of four binary input-output pairs, we add an additional feature even\footnote{For the XOR and other logical operations, perfect classification was achieved with $n=2$ using only the original features, without requiring any additional dimensions.} for this problem. This data is then normalized using MinMax scalar.

Each training input was encoded using the Baker's map with a segment length of $n=3$, and class-wise symbolic distributions were constructed accordingly. Each data instance binarized using the threshold = 0.30 and classified using the backward iteration method. In cases of ambiguity, cosine similarity between the test distribution and each class distribution was used to resolve the prediction.

\begin{figure}
    \centering
    \includegraphics[width=0.5\linewidth]{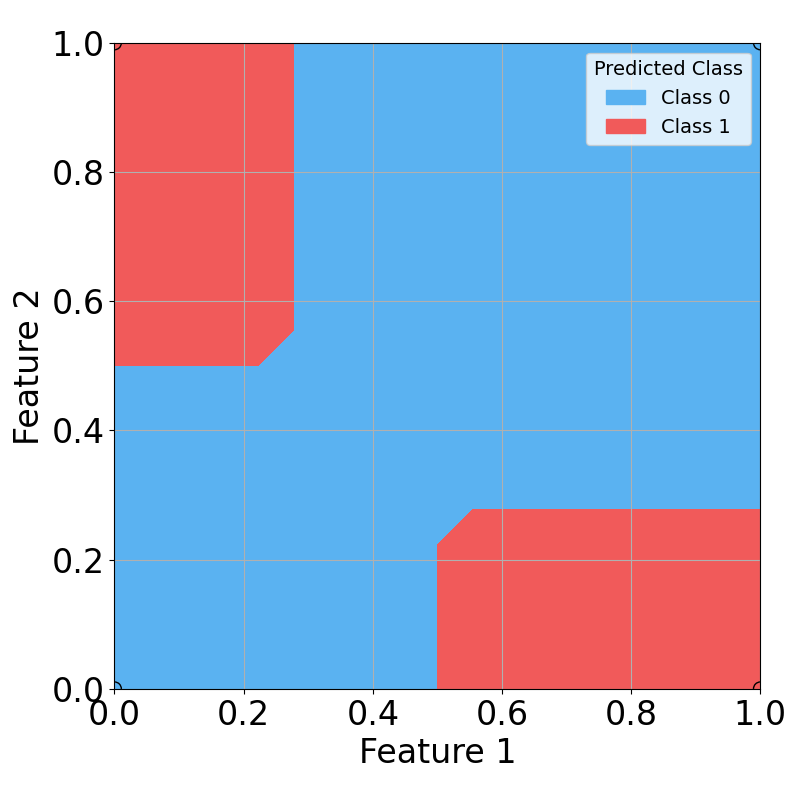}
    \caption{XOR decision boundary using \emph{ChaosComp}.}
    \label{fig:xor_problem}
\end{figure}

The classifier achieved perfect accuracy on the original XOR points. This result highlights the effectiveness of the Baker's map in capturing non-linear feature interactions and its capability to classify data that is not linearly separable. the decision boundary is shown in Figure~\ref{fig:xor_problem} Similar training and testing was carried out for NAND and NOR gates. It also achieved perfect accuracy on NAND and NOR datasets.The ability of the algorithm to solve problems like XOR, NAND, and NOR highlights the potential of chaos computing~\cite{ditto2008chaos} in handling non-linearly separable tasks.

\subsection{Experimental Setup}
Experiments were conducted on several real-world datasets, like \(Iris\)~\cite{fisher1936iris}, \(Breast Cancer\)~\cite{street1993breast} , \(Wine\)~\cite{aeberhard1991wine} , \(Seeds\)~\cite{frank2010uci} , \(Ionosphere\)~\cite{sigillito1989ionosphere} , \(Banknote\)~\cite{lohwegbanknote} , and a binary version of the \(MNIST\)~\cite{lecun1998mnist} (containing only classes 0 and 1). and synthetic datasets like circles, moons, linearly separable dataset were used from the scikit-learn library~\cite{scikit-learn}. For the \emph{Iris, Breast Cancer, Seeds, Wine, Ionosphere, and Banknote datasets, Circles, Moons, linearly separable dataset}, 80\% of the data was used for training and 20\% for testing, with a fixed random state of 90 to ensure reproducibility. Details about the split of each class's data is mentioned in Table \ref{tab:train-test_split}.
\begin{table}[!h]
\centering
  \caption{Train test split}
  \label{tab:train-test_split}
  \begin{tabular}{ccl}
    \toprule
    Dataset & Traning split & Testing split\\
    \midrule
    Iris & (41,43,36) & (9,7,14)\\
    Breast cancer & (168,287) & (44,70)\\
    Wine & (46,59,37) & (13,12,11)\\
    Seeds & (56,54,58) & (14,16,12)\\
    Ionosphere & (110,16) & (170,55)\\
    Banknote & (100,100) & (156,119)\\
    Circles & (198,202) & (52,48)\\
    Moons & (198,202) & (52,48)\\
    Linear & (198,202) & (52,48)\\
    % MNIST & (200,200) & (980,1135)\\
    
  \bottomrule
\end{tabular}
\end{table}
\begin{figure*}[!h]
  \centering
  \includegraphics[width=0.9\textwidth]{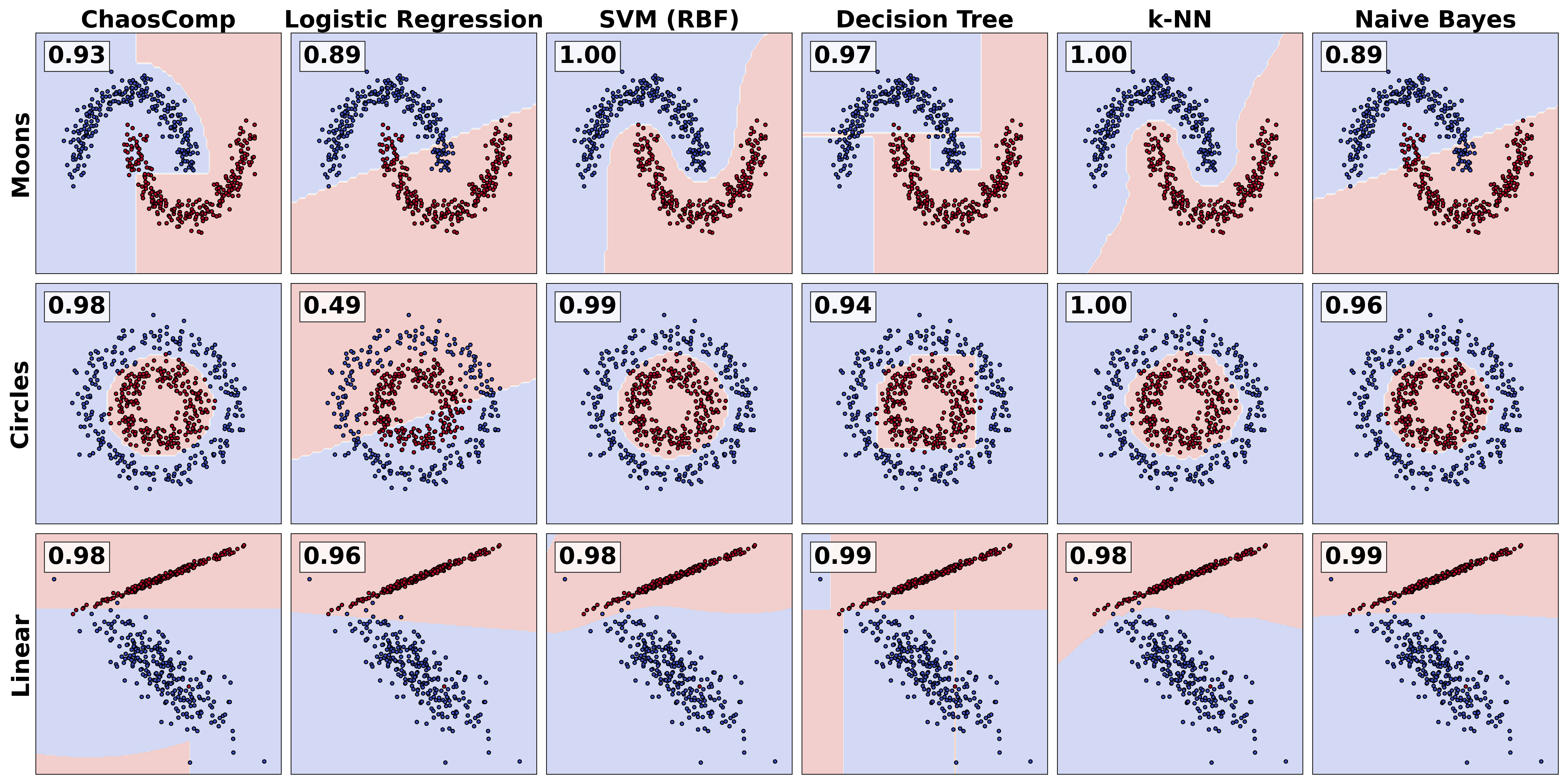}
  \caption{Test F1 scores of \emph{ChaosComp}, Logistic Regression, SVM (RBF), Decision Tree, k-NN and Naive Bayes on Circles, Moons and linearly separable dataset.}
  \label{fig:decision_boundary_models}
\end{figure*}

\begin{table*}
\centering
  \caption{Performance of \emph{ChaosComp} Results on real world datasets.}
  \label{tab:results}
  \begin{tabular}{ccccccl}
    \toprule
    Dataset & Optimal Threshold & n-value & Accuracy & Macro Precision & Macro Recall & Macro F1-score \\
    \midrule
    Iris & 0.59 & 4 & 0.8667 & 0.8788 & 0.8651 & 0.8469 \\
    Breast cancer & 0.32 & 4 & 0.9561 & 0.9604 & 0.9474 & 0.9531 \\
    Wine & 0.19 & 4 & 0.9167  & 0.9286 & 0.9167 & 0.9124  \\
    Seeds & 0.46 & 4 & 0.9524  & 0.9583 & 0.9444 & 0.9475  \\
    Ionosphere & 0.52 & 4 & 0.8169  & 0.7414 & 0.7710 & 0.7535  \\
    Banknote & 0.58 & 4 & 0.8982  & 0.8971 & 0.9043 & 0.8976  \\
    % Iris (PCA) (3 features) & 0.19 & 4 & 0.9000 & 0.9003 & 0.9153 & 0.9017 \\
    % Breast cancer (PCA) (2 features) & 0.32 & 2  & 0.9737 & 0.9795 & 0.9659 & 0.9719 \\
    % Wine (PCA) (3 features) & 0.46 & 2 & 0.9722  & 0.9722 & 0.9722 & 0.9710  \\
    % Seeds (PCA) (4 features) & 0.48 & 4 & 0.8810  & 0.8804 & 0.8690 & 0.8718  \\
    % Ionosphere (PCA) (3 features) & 0.60 & 3 & 0.9155  & 0.8935 & 0.8568 & 0.8732  \\
    % Banknote (PCA) (4 features) & 0.44 & 4 & 0.8727  & 0.8720 & 0.8789 & 0.8721  \\
    % Binary-MNIST (PCA) (2 features) & 0.66 & 3 & 0.9953 & 0.9955 & 0.9950  & 0.9952 \\
    \bottomrule
  \end{tabular}
\end{table*}
To reduce computational overhead, the number of training samples per class was limited to  100 for the Banknote dataset, while the full test sets were retained for evaluation. The optimal threshold and \( n \)-value for the \( n \)-th return map were selected via five-fold cross-validation on all datasets.

\subsection{Results}

In Experiment 1 on synthetic datasets, decision boundaries along with macro F1 scores are shown in Figure~\ref{fig:decision_boundary_models}. For Experiments 2, the performance of \emph{ChaosComp} is summarized in Table~\ref{tab:results}, including the selected values of \( n \), threshold, and \( n\_components \) (determined via cross-validation). A fixed Laplace smoothing parameter of \( \alpha = 0.01 \) was used in all experiments to address zero-frequency symbolic sequences during training.

\begin{table*}
\centering
  \caption{Comparison performance analysis of \emph{ChaosComp} with Logistic Regression, SVM (RBF), Decision Tree, k-NN and Naive Bayes. The performance measure used for comparative study is test Macro F1 score.}
  \label{tab:comparison}
  \begin{tabular}{ccccccl}
    \toprule
    Model & Iris & Wine & Breast cancer & Seeds & Ionosphere & Banknote \\
    \midrule
    \emph{ChaosComp} & 0.8469 & 0.9124 & 0.9531 & 0.9475 & 0.7535 & 0.8976 \\
    Logistic Regression & 0.9654 & 0.9710 & 0.9526 & 0.9484 & 0.8221 & 0.9926 \\
    KNN & 0.9654 & 0.9429 & 0.9526  & 0.9743 & 0.7942 & 1.0000  \\
    SVM & 0.9654 & 0.9710 & 0.9531  & 0.9484 & 0.8790 & 1.0000  \\
    Decision tree & 0.9654 & 0.8618 & 0.9260  & 0.8701 & 0.7829 & 0.9889  \\
    Naive Bayes & 0.9654 & 0.9177 & 0.9526  & 1.0000 & 0.8221 & 0.8801  \\
    \bottomrule
  \end{tabular}
\end{table*}
% If your title is lengthy, you must define a short version to be used
% in the page headers, to prevent overlapping text. The \verb|title|
% command has a ``short title'' parameter:
% \begin{verbatim}
%   \title[short title]{full title}
% \end{verbatim}
% \begin{table*}
% \centering
%   \caption{Comparison study of \emph{ChaosComp} with different models with PCA.}
%   \label{tab:comparison_pca}
%   \begin{tabular}{cccccccl}
%     \toprule
%     Model & Iris & Wine & Breast cancer & Seeds & Ionosphere & Banknote & Binary-MNIST \\
%     \midrule
%     \emph{ChaosComp} & 0.9017 (3) & 0.9710 (3) & 0.9719 (2) & 0.8718 (4) & 0.8732 (3) & 0.8721 (4) & 0.9953 (2)\\
%     Logistic Regression & 0.9654 (3) & 0.9444 (2) & 0.9627 (2) & 0.9743 (4) & 0.8224 (3) & 1.0000 (4) & 0.9952 (4) \\
%     KNN & 1.0000 (2) & 0.9710 (2) & 0.9531 (2)  & 0.9743 (4) & 0.8556 (3) & 1.0000 (4) & 0.9967 (4) \\
%     SVM & 0.9654 (2) & 0.9177 (3) & 0.9627 (2)  & 0.9484 (2) & 0.8483 (3) & 1.0000 (4) & 0.9962 (2) \\
%     Decision tree & 0.9654 (3) & 0.9444 (2) & 0.9343 (4)  & 0.8805 (4) & 0.8512 (3) & 0.9852 (4) & 0.9943 (4)  \\
%     Naive Bayes & 0.9654 (2) & 0.9444 (2) & 0.9623 (3)  & 1.0000 (4) & 0.7646 (4) & 0.9633 (4) & 0.9967 (2) \\
%     \bottomrule
%   \end{tabular}
% \end{table*}

\subsection{Comparitive study with ML models}
We conducted a comparative study of \emph{ChaosComp} with other standard ML models like Logistic Regression, SVM, Decision tree, k-NN, and Naive Bayes on synthetic as well as real world datasets. Hyperparameter tuning was done for each of these models using five fold cross validation and gridsearch was used to find the best hyperparameters. Results for synthetic datasets is mentioned in the figure~\ref{fig:decision_boundary_models}, while table~\ref{tab:comparison} contains results on real world datasets. \emph{ChaosComp} achieved the highest F1 score of 0.9531 on the Breast Cancer dataset and delivered strong performance on the Seeds dataset with an F1 score of 0.9475, outperforming Decision Tree. 
% Table~\ref{tab:comparison_pca} shows the comparison study when PCA was used as a preprocessing step. \emph{ChaosComp} achieved the highest F1 score of 0.9710 on the Wine dataset, 0.9719 on the Breast Cancer dataset, 0.8732 on the Ionosphere dataset and 0.9967 on the Binary-MNIST dataset. The decision boundary for Breast Cancer dataset when PCA was used is mentioned in Figure~\ref{fig:bsw_decision}.

\section{Limitations}
The proposed \emph{ChaosComp} classifier shows promising performance. However, several limitations offer opportunities for further exploration. The backward iteration method faces precision issues when there are a lot of features. This issue can be resolved using the renormalization~\cite{sayood2017introduction} during backiteration phases. Introducing re-normalization steps during the backiteration phase improves numerical stability and convergence. \section{Future Work}
An interesting direction is to explore the theoretical link between the \emph{learnability} (the minimum number of bits needed to represent a class in a way that allows it to be reliably distinguished from other classes) of a class and the \emph{degree of chaos} in its associated \( n \)-th return map. 

For the standard Baker's's map, the Lyapunov exponent equals the Shannon entropy:
\[
\lambda = -a \log a - (1 - a) \log(1 - a).
\]

For the \( n \)-th return map, constructed from empirical probabilities \( \{p_w\}_{w \in \{0,1\}^n} \), the entropy generalizes to:
\[
H_n = -\sum_{w \in \{0,1\}^n} p_w \log p_w.
\]

This entropy corresponds to the \emph{average code length} or \emph{compressed size}, linking symbolic dynamics, chaos, and information content. Studying how \( H_n \) correlates with class separability may offer foundational insights into compression-based learning.

\section{Conclusion}
In this paper, we propose \emph{ChaosComp}, a classification algorithm founded on the principles of symbolic dynamics and compression using chaotic maps. Each class is modeled as a distinct chaotic dynamical system through empirical symbolic transition probabilities derived from training data. Classification is performed by selecting the class that yields the shortest compressed representation of a test instance, as measured via backward iteration on class-specific return maps. The proposed method has been evaluated on both simulated and real world datasets, demonstrating promising performance (test macro F1 scores for Breast Cancer = 0.9531, Seeds = 0.9475, Wine = 0.9124, Banknote = 0.8976, Iris = 0.8469, Ionosphere = 0.7535)  while requiring only two hyperparameters. Beyond empirical results, this work lays the foundation for a deeper theoretical investigation into the relationship between chaos, entropy, and learnability. In particular, future research may explore how the entropy of class-specific return maps affects classification performance.
\section*{Code Availability}
The source code supporting this research is publicly available at the following GitHub repository: \url{https://github.com/i-to-the-power-i/ChaosComp}.
\section*{Acknowledgments}
Harikrishnan N. B. gratefully acknowledges the financial support from the Prime Minister's Early Career Research Grant (Project No. ANRF/ECRG/2024/004227/ENS). The authors also express their sincere thanks to Prof. Nithin Nagaraj for his valuable discussions on the relationship between Shannon entropy and the Lyapunov exponent of the Baker's map.
\bibliography{reference}

\end{document}